\documentclass[letterpaper, 10 pt, conference]{ieeeconf}  
\IEEEoverridecommandlockouts                              
\usepackage{amsmath,amsfonts}
\usepackage{algorithmic}
\usepackage{algorithm}
\usepackage{array}

\usepackage[caption=false,font=normalsize,labelfont=sf,textfont=sf]{subfig}

\usepackage{textcomp}
\usepackage{stfloats}
\usepackage{url}
\usepackage{verbatim}
\usepackage{graphicx}

\usepackage{multirow} 
\usepackage{pifont} 
\usepackage{colortbl} 
\usepackage{xcolor}

\usepackage{hyperref}
\hypersetup{
	colorlinks=true,
	linkcolor=cyan,
	filecolor=blue,      
	urlcolor=black,
	citecolor=green,
}

\title{\LARGE \bf
FisheyeDepth: A Real Scale Self-Supervised Depth Estimation Model for Fisheye Camera
}
\author{Guoyang Zhao, Yuxuan Liu, Weiqing Qi, Fulong Ma, Ming Liu, and Jun Ma, \textit{Senior Member, IEEE} 
\thanks{G. Zhao, W. Qi, F. Ma, M. Liu, and J. Ma are with the Robotics and Autonomous Systems Thrust, The Hong Kong University of Science and Technology (Guangzhou), Guangzhou, China
(e-mail:{\{gzhao492, wqiad, fmaaf\}@connect.hkust-gz.edu.cn; eelium@hkust-gz.edu.cn; jun.ma@ust.hk).} \textit{(Corresponding author: Jun Ma.)}}
\thanks{Y. Liu is with the Department of Electronic and Computer Engineering, The Hong Kong University of Science and Technology, Hong Kong SAR, China
(e-mail: yliuhb@connect.ust.hk).}
}

\begin{document}

\maketitle
\thispagestyle{empty}
\pagestyle{empty}

\begin{abstract}

Accurate depth estimation is crucial for 3D scene comprehension in robotics and autonomous vehicles. Fisheye cameras, known for their wide field of view, have inherent geometric benefits. However, their use in depth estimation is restricted by a scarcity of ground truth data and image distortions.
We present FisheyeDepth, a self-supervised depth estimation model tailored for fisheye cameras. 
We incorporate a fisheye camera model into the projection and reprojection stages during training to handle image distortions, thereby improving depth estimation accuracy and training stability.
Furthermore, we incorporate real-scale pose information into the geometric projection between consecutive frames, replacing the poses estimated by the conventional pose network.
Essentially, this method offers the necessary physical depth for robotic tasks, and also streamlines the training and inference procedures. Additionally, we devise a multi-channel output strategy to improve robustness by adaptively fusing features at various scales, which reduces the noise from real pose data. We demonstrate the superior performance and robustness of our model in fisheye image depth estimation through evaluations on public datasets and real-world scenarios.
The project website is available at: \href{https://github.com/guoyangzhao/FisheyeDepth}{https://github.com/guoyangzhao/FisheyeDepth}. 

\end{abstract}

\section{INTRODUCTION}

Depth estimation plays a crucial role in 3D scene perception in the fields of robotics and autonomous driving. Due to the sparse nature and high cost of LiDAR point cloud-based depth sensing, image-based perception methods hold significant value in terms of coverage density and redundancy \cite{kumar2018monocular, zhao2024tsclip}. Fisheye cameras, with their wide field of view (FOV) and advanced geometric properties, have seen a notable increase in use for perception tasks in recent years \cite{yogamani2024fisheyebevseg,liu2024omnicolor}. However, the majority of research have focused on pinhole cameras for depth estimation \cite{zhou2017unsupervised,garg2016unsupervised,godard2019digging,mahjourian2018unsupervised}, while research dedicated to fisheye cameras remains relatively scarce \cite{zioulis2018omnidepth}.

Researchers have conducted extensive studies on depth estimation using learning-based methods with narrow FOV cameras \cite{bae2023deep,zhang2023lite,sun2023sc}, but these efforts primarily focus on traditional 2D content captured by cameras, adhering to the typical pinhole projection model based on rectified image sequences. With the rise of affordable wide-angle fisheye cameras, depth estimation has been extended to omnidirectional content through methods like omnidirectional stereo \cite{jiang2024romnistereo,chen2023s} and Structure from Motion (SfM) \cite{scaramuzza2006flexible,kim2024omnisdf}. However, these approaches often rely on LiDAR or high-precision maps to generate dense, pixel-wise labels for ground truth data, such as in the KITTI dataset \cite{geiger2013vision}. This necessitates multiple LiDAR sensors to compensate for blind spots at close ranges, making the overall setup both expensive and time-consuming, thus limiting the data available for model training \cite{zhao2024curbnet}. Furthermore, when projecting LiDAR point clouds into pixel-wise distance labels for fisheye cameras, significant motion distortions often occur.


Self-supervised methods \cite{zhan2018unsupervised,xue2020toward,zhao2023gasmono} alleviate the challenge of acquiring ground truth data by training neural networks to optimize depth maps based on the geometric projection parameters between stereo images or consecutive frames. However, these approaches are susceptible to issues such as occlusion, blurriness, and varying lighting conditions between consecutive frames, which can lead to inaccurate depth predictions during self-supervised optimization \cite{godard2019digging}. Moreover, fisheye camera images exhibit significant distortion, posing severe challenges for pixel-wise geometric transformation between consecutive images.

Several studies have attempted to perform depth estimation for fisheye images within a self-supervised framework. \cite{kumar2020fisheyedistancenet,kumar2021fisheyedistancenet++} leverage convolutional neural networks to regress Euclidean distance maps directly from raw fisheye image sequences, utilizing self-attention encoding and specially designed loss functions to produce sharp depth maps. \cite{kumar2021syndistnet} proposes a joint learning framework for self-supervised distance estimation and semantic segmentation to mitigate the interference of dynamic artifacts between consecutive frames. While these methods aim to alleviate fisheye image distortion by improving model architectures, they do not fully resolve the geometric transformation issues inherent to fisheye cameras, particularly in regions with severe distortion near the image edges. Additionally, these models do not account for accurate real motion scale during training, limiting their applicability in real-world robotic systems.


We propose a real scale self-supervised depth estimation model for fisheye cameras, tailored for robotic scenarios. The model is built upon the Monodepth2 \cite{godard2019digging} depth estimation network with several key modifications.
First, we apply a fisheye camera model to project the estimated depth, accurately handling the distortion inherent in fisheye images and ensuring stable model training.
Second, during the geometric projection between consecutive frames, we introduce real scale poses to replace the poses estimated by the original pose network, assisting the self-supervised depth optimization. This not only meets the requirement for physical depth in robotic applications but also simplifies the complexity of both training and inference.
Finally, we design a multi-channel output module, which enhances feature fusion at multiple stages to mitigate noise introduced by real-world pose data, thereby improving the robustness of depth estimation for fisheye images.
In summary, our main contributions include:

\begin{itemize}
\item[1)] We propose a fisheye projection scheme based on the camera model to eliminate distortion in training, which greatly improves the accuracy of depth estimation.
\item[2)] We incorporate real scale poses during the training process, which renders the model suitable for real-world robotic interactions and navigation tasks.
\item[3)] We devise a multi-channel output module that performs adaptive feature fusion across multiple scales, and this ensures robust depth predictions across different scenes.
\item[4)] As illustrated in the comparative experiments, the FisheyeDepth model surpasses traditional self-supervised models and achieves stable self-supervised depth estimation on fisheye images with full-scale distortion for the first time.
\end{itemize}


\section{RELATED WORKS}

\subsection{Monocular Self-Supervised Depth Estimation}
Monocular self-supervised depth estimation typically involves training neural networks by optimizing depth maps as geometric projection parameters between stereo or consecutive images \cite{zhou2017unsupervised}. For instance, \cite{xue2020toward,garg2016unsupervised} model depth as part of the geometric projection between stereo and sequential images, enabling self-supervised network training. The initial self-supervised approaches have since been extended to various depth estimation strategies. For example, \cite{aleotti2018generative,gordon2019depth} introduced improvements in loss functions, while \cite{liu2024degan,ke2024repurposing} employed Generative Adversarial Networks to optimize self-supervised depth estimation. These methods often rely on proxy labels generated from traditional stereo algorithms \cite{tosi2019learning} or synthetic data \cite{tabata2023analyzing} for further expansion.

Several approaches have also developed dedicated self-supervised depth estimation frameworks \cite{han2023self,liu2023self}, such as leveraging teacher-student learning models for training \cite{liu2024estimating}, with refined depth estimation strategies applied during testing \cite{casser2019depth,liu2023fsnet}. Moreover, some methods predict camera parameters \cite{gordon2019depth,chanduri2021camlessmonodepth} to facilitate cross-camera training using images captured by different cameras. While all these methods have demonstrated remarkable success on pinhole camera images, they have rarely been adapted to more complex camera models, such as fisheye cameras \cite{xie2023omnividar}.

\subsection{Fisheye Self-Supervised Depth Estimation}
Due to the complex geometric properties of fisheye cameras, self-supervised depth estimation methods have not been extensively studied in this domain. \cite{kumar2020fisheyedistancenet,kumar2020unrectdepthnet} were the first to demonstrate the applicability of self-supervised depth estimation for fisheye camera distance estimation tasks. Earlier works \cite{wang2020bifuse,jin2020geometric} also explored the use of self-supervised depth estimation in 360° images.

Methods such as \cite{kumar2020fisheyedistancenet,kumar2021fisheyedistancenet++,kumar2021syndistnet,kumar2021svdistnet} proposed self-supervised depth estimation frameworks for fisheye images. \cite{kumar2020fisheyedistancenet} attempted to learn Euclidean distance and self-motion from raw monocular fisheye videos, but encountered issues with sampling distortion and discontinuities in transitional regions. \cite{kumar2021fisheyedistancenet++} introduced self-attention layers and robust loss functions to improve depth map clarity and accelerate training convergence. \cite{kumar2021syndistnet} proposed a joint framework for semantic segmentation and self-supervised depth estimation, guiding depth estimation through accurate semantic features. \cite{kumar2021svdistnet} designed an adaptive multi-scale network to enhance the generalization of fisheye self-supervised depth estimation, improving performance through self-attention encoding.

Although these methods mitigate some fisheye image distortion through improved models and frameworks, they do not fully address the geometric transformation challenges \cite{lee2023slabins}. Severe distortion, especially at the image edges, still affects depth estimation. Additionally, the lack of real motion scale consideration during training limits their use in real-world robotic applications.

\subsection{Fisheye Depth Estimation Dataset}
In the field of robotics, fisheye cameras are gaining attention due to their wide FOV. However, publicly available datasets for fisheye images remain relatively scarce, especially for outdoor scenes of autonomous driving.

The KITTI-360 \cite{liao2022kitti} dataset is one of the few comprehensive fisheye image datasets that is publicly available. It provides rich sensor data and detailed annotations, including dense depth, semantic, and instance annotations for both 3D point clouds and 2D images. The dataset was collected using 180° fisheye cameras mounted on both sides of the vehicle, along with a top-mounted push-broom LiDAR scanner, enabling a complete 360° view, making it a valuable resource for fisheye depth estimation tasks.

The WoodScape \cite{yogamani2019woodscape} dataset is designed to advance the application of fisheye cameras in low-speed vehicle scenarios. It features four surround-view cameras and provides instance-level semantic annotations for over 10,000 images, covering nine tasks including segmentation, depth estimation, and 3D bounding box detection. However, WoodScape does not public the ground truth of fisheye image depth, which is a limitation for quantitative evaluation.

\section{METHODOLOGY}

\subsection{Framework}

\begin{figure*}[t!]
    \centering
    \includegraphics[width=.96\textwidth]{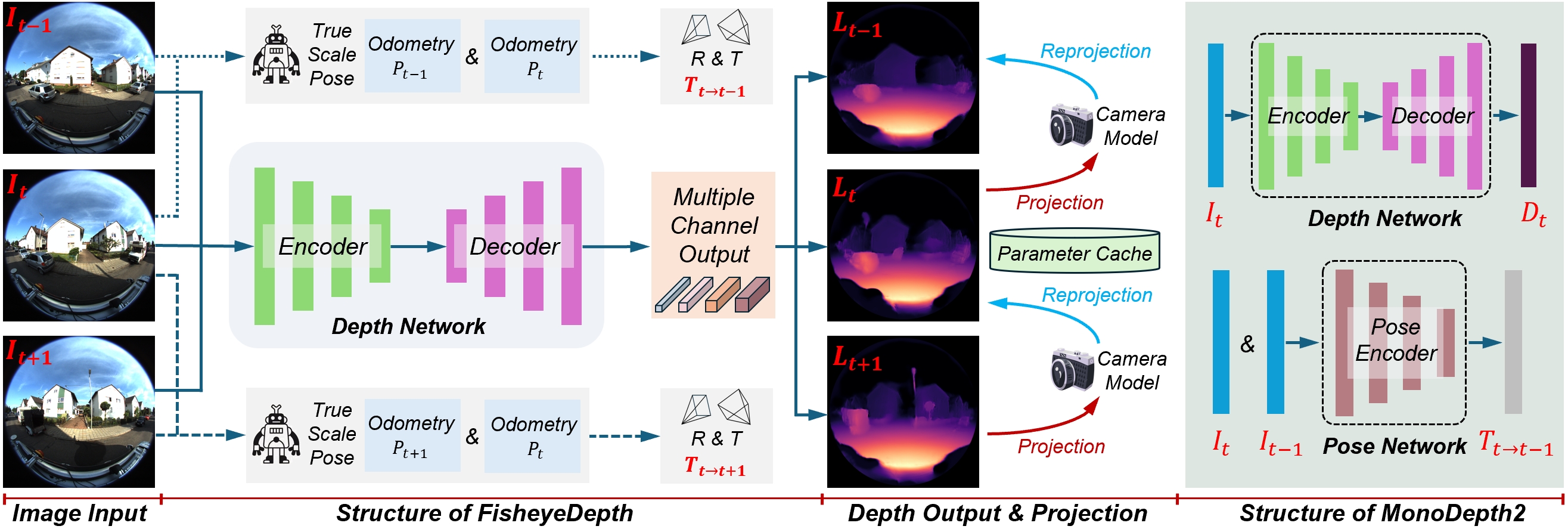}
    \vspace{-5pt}

    \caption{\textbf{Structure of the FisheyeDepth model.} (1) We introduce a fisheye camera model during training to reduce projection distortion. (2) Real-scale poses from the robot are incorporated into the training process. (3) A multi-channel output is proposed to ensure stable training through feature fusion.}
    \label{framework}
    \vspace{-15pt}
\end{figure*}

The structure of our proposed FisheyeDepth model is illustrated in Fig. \ref{framework}. The model primarily takes fisheye images and their adjacent frames as input for depth estimation, utilizing consecutive frames to compute relative poses. To facilitate application across various scenes, we build upon the established Monodepth2 model, which is a self-supervised depth estimation method designed for monocular camera images. our model mainly retains the encoder and decoder from the depth network for feature extraction and depth estimation, where the encoder employs a pre-trained ResNet18 model.


Differing from the original model, we adapt the self-supervised depth estimation framework to address the characteristics of fisheye images from the perspective of practical robotic applications. First, we introduce a fisheye camera model in the depth output section to handle projection and reprojection, addressing the distortion inherent in fisheye images. Accurate distortion modeling ensures stable convergence during model training. Second, we remove the original PoseNet and directly use real-scale pose information from the robot for rotation and translation calculations. Thirdly, we modify the network to output a distance map $L=\sqrt{X^2 + Y^2 + Z^2}$ instead of depth map $Z$ which provides more stable and continuous values that align better with the geometric properties of the fisheye projection model. Finally, we add a multi-scale output module at the end of the model, enhancing feature fusion to mitigate the noise introduced by real-scale poses, thereby improving the robustness of depth estimation across different scenes.

\subsection{Flash Fisheye Projection}
In conventional self-supervised depth estimation methods like Monodepth2, the projection and reprojection processes during network training are typically handled with the assumption of negligible camera distortion, using a pinhole camera model. However, this assumption does not hold for fisheye images, which exhibit significant radial distortion. Directly applying standard projection techniques to fisheye images can lead to large errors. To address this, we propose a projection and reprojection scheme tailored for fisheye images, based on the Mei unified camera model~\cite{mei2007single}. By accounting for fisheye camera distortion during the projection process, we ensure accurate pixel correspondences during training, which is crucial for calculating the photometric loss and achieving stable network convergence.


\subsubsection{\textbf{Projection Process}}
For the depth predictions output by the decoder, we first project the 3D points from the camera coordinate system onto the fisheye image plane. Given a 3D point $\mathbf{X} = (X, Y, Z)^\top$ in the camera coordinate system, the projection is performed in the following steps. First, we normalize the point onto the unit sphere as $\mathbf{X}' = \frac{\mathbf{X}}{\|\mathbf{X}\|}$. Then, using the mirror parameter $\xi$, the point is projected onto a normalized plane:
\begin{equation}
x = \frac{X'}{Z' + \xi}, \quad y = \frac{Y'}{Z' + \xi}.
\end{equation}
To handle the radial distortion of the fisheye image, we apply the radial distortion model to correct the coordinates:
\begin{equation}
x_d = x \left(1 + k_1 r^2 + k_2 r^4\right), \quad y_d = y \left(1 + k_1 r^2 + k_2 r^4\right),
\end{equation}
where $r^2 = x^2 + y^2$, and $k_1$ and $k_2$ are distortion coefficients. Finally, the normalized coordinates are mapped to pixel coordinates using the camera's intrinsic matrix:
\begin{equation}
u = \gamma_1 x_d + u_0, \quad v = \gamma_2 y_d + v_0,
\end{equation}
where $\gamma_1$ and $\gamma_2$ are focal length scaling factors, and $(u_0, v_0)$ are the principal point coordinates.


\subsubsection{\textbf{Reprojection Process}}
For self-supervised training, the photometric loss is computed by reprojection, where pixels from the source image are reprojected onto the target image plane. This process involves reversing the steps of the projection. First, pixel coordinates are converted back to normalized coordinates:
\begin{equation}
x_d = \frac{u - u_0}{\gamma_1}, \quad y_d = \frac{v - v_0}{\gamma_2}.
\end{equation}

Next, the radial distortion is removed using the Newton-Raphson method \cite{ypma1995historical}. The radial distance $r$ is solved iteratively using $r = \text{NewtonRaphson}(r_d)$, where $r_d = \sqrt{x_d^2 + y_d^2}$. The undistorted coordinates are then given by:
\begin{equation}
x = \frac{x_d}{1 + k_1 r^2 + k_2 r^4}, \quad y = \frac{y_d}{1 + k_1 r^2 + k_2 r^4},
\end{equation}

Finally, the inverse mirror model is used to compute $Z'$, and the 3D point $\mathbf{X}$ in the camera coordinates is recovered by the predicted depth $D$, as $\mathbf{X} = D \cdot \mathbf{X}'$.

\subsubsection{\textbf{Iterative Solvers}}
Due to the non-linear nature of inverse radial distortion correction and the mirror model reprojection, we employ iterative numerical methods. Newton-Raphson is used for inverting the radial distortion, through:
\begin{equation}
r_{n+1} = r_n - \frac{f(r_n)}{f'(r_n)},
\end{equation}
where $f(r) = r (1 + k_1 r^2 + k_2 r^4) - r_d$. Additionally, the bisection method is applied to solve the inverse mirror model, ensuring convergence by narrowing the solution interval.


To address the substantial computational overhead caused by the extensive projection calculations during training, we implement a caching mechanism. After the first training iteration, we store the projection parameters for each camera, significantly reducing the need for repeated camera projection calculations. By incorporating this fisheye-specific projection scheme, we enhance depth estimation accuracy and ensure stable training with more reliable convergence.

\subsection{Training with Real-Scale Pose}

In practical robotic applications, accurate depth estimation is crucial for navigation and interaction with the environment. Traditional self-supervised methods, such as Monodepth2, rely on a PoseNet to estimate relative poses from adjacent image frames. However, these pose estimates lack real scale due to their sole reliance on visual information, leading to scale ambiguity in depth predictions. This not only increases the complexity of model training and inference but also limits the applicability in real-world scenarios.

To address these, we eliminate the PoseNet from Monodepth2 and instead directly employ real-scale pose information obtained through sensor fusion of IMU, LiDAR, and Camera data. By integrating odometry from these sensors, we acquire accurate real-world poses that inherently contain true scale. This approach simplifies the model architecture by removing the need for learning pose estimation and enhances the depth prediction by the scale-aware pose information.

During training, for a given current frame $I_t$ and its adjacent frames $I_{t'}$, we utilize the real relative poses $\mathbf{T}_{t \rightarrow t'}$ obtained from sensor fusion. The depth estimation network first extracts features from $I_t$ and predicts the depth map $\mathbf{L}_t$. Using the real-scale poses, we project the pixels from the current frame to the adjacent frames, which facilitates the computation of the photometric reconstruction loss essential for self-supervised training.

The pixel mapping process is defined as follows: for a pixel coordinate $\mathbf{p}_t$ with distance $L_t(\mathbf{p}_t)$ in the current frame, we compute its corresponding coordinate $\mathbf{p}_{t'}$ in the adjacent frame using the fisheye projection function $\mathbf{F}$, the re-projection function $\mathbf{F}^{-1}$ and the real pose $\mathbf{T}_{t \rightarrow t'}$:
\begin{equation}
\mathbf{p}_{t'} \sim \mathbf{F} (\mathbf{T}_{t \rightarrow t'}  \mathbf{F}^{-1} (\tilde{\mathbf{p}}_t)  )
\end{equation}
where $\tilde{\mathbf{p}}_t$ is the homogeneous coordinate of $\mathbf{p}_t$, and $\sim$ denotes equality up to a scale.

By leveraging real-scale poses, we avoid the scale ambiguity inherent in pose estimates from PoseNet, leading to more accurate depth predictions. Additionally, this approach allows for the joint training of multiple cameras, as we can obtain the real scale and relative poses for all cameras involved. 

\subsection{Multi-Channel Output}

This module outputs depth predictions at different stages of the decoder, combines adaptive feature fusion using a channel attention mechanism, optimizes the noise introduced by real-scale poses. 

In the decoder, we adopt a strategy of layer-by-layer upsampling and feature fusion. At each decoding stage $i$, the decoder generates a depth prediction output. Specifically, for the feature map $X_i$ at the $i$-th layer of the decoder, we introduce a channel attention mechanism to achieve adaptive feature fusion.

First, we apply the channel attention mechanism to the feature map $X_i$ to compute the attention weights $A_i$:
\begin{equation}
A_i = \sigma(\text{Conv}_{\text{attn}}(X_i))
\end{equation}
where $\text{Conv}_{\text{attn}}$ denotes the convolutional layer used to compute the attention weights, and $\sigma$ is the sigmoid activation function. Then, we weight the feature map $X_i$:
\begin{equation}
\tilde{X}_i = A_i \odot X_i
\end{equation}
where $\odot$ denotes element-wise multiplication. Through the channel attention mechanism, the model can automatically learn the importance of different channel features, highlight key features, and achieve adaptive feature fusion.

Next, we use a convolutional layer to map the weighted feature map $\tilde{X}_i$ to the depth prediction logits $\text{logits}_i$:
\begin{equation}
\text{logits}_i = \text{Conv}_{\text{disp}, i}(\tilde{X}_i)
\end{equation}

Then, using the $\text{gather\_output}$ function, we convert $\text{logits}_i$ into the actual distance map $L_i$ and disparity map $\text{disp}_i$:
\begin{equation}
L_i,\, \text{disp}_i = \text{gather\_output}(\text{logits}_i,\, \text{depth\_scale})
\end{equation}
where $\text{depth\_scale}$ is a scaling factor computed based on the camera's intrinsic parameters. The $\text{gather\_output}$ function maps the model's output from the high-dimensional feature space to the actual depth value range.

By outputting multi-scale depth predictions at different stages of the decoder and combining adaptive feature fusion using the channel attention mechanism, the model can capture depth information at different resolutions. This multi-channel output approach helps to smooth the noise introduced by real-scale poses, enhancing the accuracy and robustness of depth estimation.






\begin{table*}[t]
\renewcommand\arraystretch{1.45}
\caption{Quantization experimental results in the KITTI-360 dataset. (Real Scale means using the robot’s pose for training)}
\vspace{-8pt}
\centering
\footnotesize
\begin{tabular}{>{\centering\arraybackslash}m{1.8cm}cccccccccc}
\hline 
Camera Model & Methods & Real Scale & \cellcolor{cyan!30}Abs Rel\(\downarrow\) & \cellcolor{cyan!30}Sq Rel\(\downarrow\) & \cellcolor{cyan!30}RMSE\(\downarrow\) & \cellcolor{cyan!30}RMSE$_{\log }$\(\downarrow\) & \cellcolor{pink!30}$\delta<1.25$ \(\uparrow\) & \cellcolor{pink!30}$\delta<1.25^2$ \(\uparrow\) & \cellcolor{pink!30}$\delta<1.25^3$ \(\uparrow\) \\
\hline 
\multirow{4}{*}{\bf{Monocular}} 
& Monodepth2 \cite{godard2019digging} & \textcolor{red}{\ding{55}} & 0.602 & 1.804 & 1.788 &  0.563 & 0.426 & 0.685 & 0.835 \\
& DNet \cite{xue2020toward} & \textcolor{red}{\ding{55}} & 0.558 & 1.746 & 1.802 & 0.560 & 0.443 & 0.691 & 0.838 \\
& FSNet \cite{liu2023fsnet} & \textcolor{red}{\ding{55}} & 0.464 & 1.117 & 1.125 & 0.339 & 0.613 & 0.768 & 0.868 \\
& FSNet \cite{liu2023fsnet} & \textcolor{green}{\ding{51}} & 0.435 & 1.012 & 0.937 & 0.309 & 0.633 & 0.792 & 0.871 \\
\hline
\multirow{1}{*}{\bf{Fisheye}} 
& FisheyeDepth & \textcolor{green}{\ding{51}} & \textbf{0.106} & \textbf{0.407} & \textbf{0.927} & \textbf{0.170} & \textbf{0.920} & \textbf{0.975} & \textbf{0.989}\\
\hline
\end{tabular}
\label{Table1}
\vspace{-5pt}
\end{table*}

\begin{figure*}[t!]
    \centering
    \includegraphics[width=.99\textwidth]{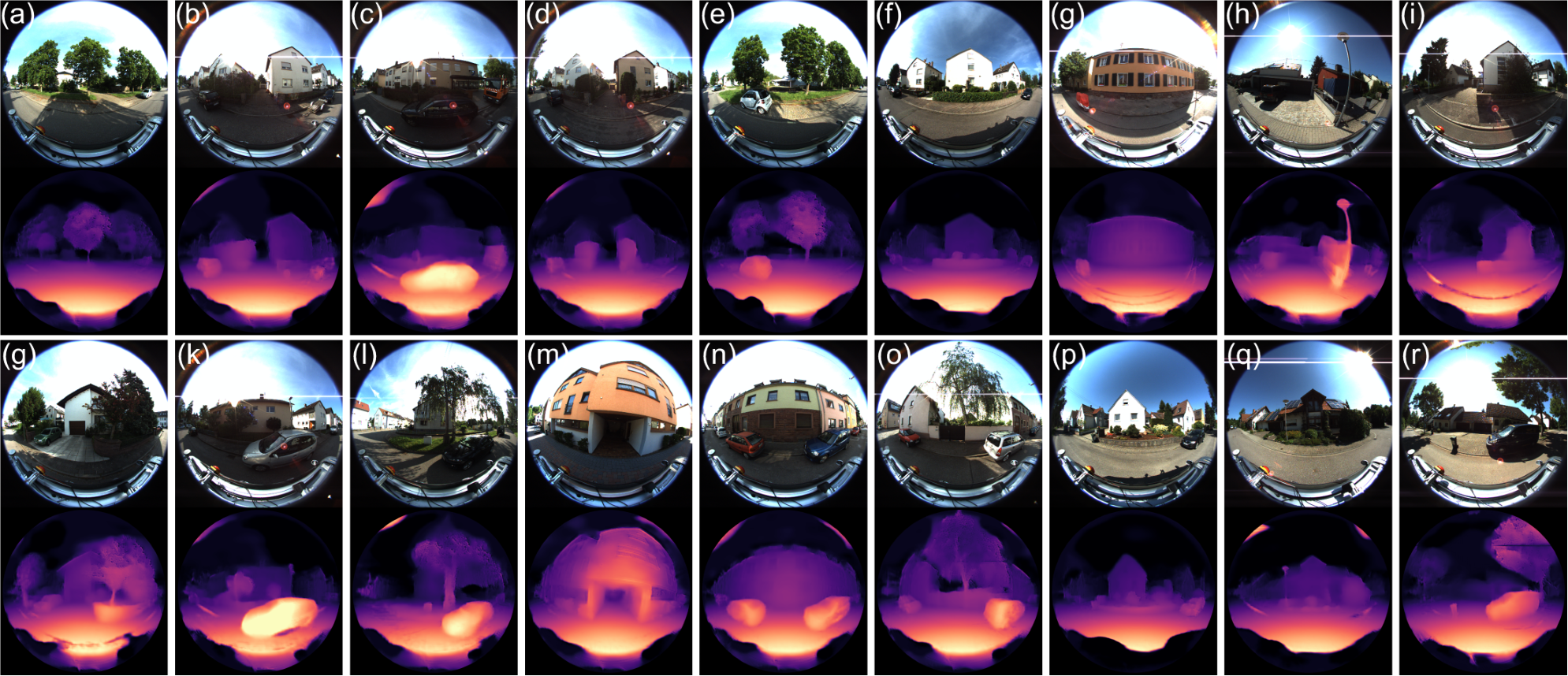}
    \vspace{-8pt}
    \caption{\textbf{Depth estimation visualization in the KITTI-360 dataset.} The image background contains various elements such as roads, vegetation, buildings, and cars in urban road scenes.}
    \label{result360}
    \vspace{-15pt}
\end{figure*}

\section{EXPERIMENTS}

\subsection{Experiment Setup}
Our model training was conducted in an Ubuntu 20.04 environment, utilizing an Intel(R) Xeon(R) Gold 5318S CPU @ 2.10GHz and an NVIDIA RTX 3090 GPU. We employed the PyTorch framework for model training and set training parameters with a batch size of 8, a total of 20 epochs, and a learning rate of 1e-4. To prevent dynamic objects from affecting the depth prediction during training, we mask the vehicle body in the input images.

\subsection{Quantitative Results in KITTI-360}

Table \ref{Table1} presents the quantitative results in the KITTI-360 dataset. In KITTI-360, we jointly train on images from two fisheye cameras, with ground-truth poses obtained from LiDAR inertial odometry.

We primarily compare our method with the self-supervised monocular depth estimation models Monodepth2 and DNet, which do not utilize real-scale information, and the FSNet model, which incorporates real-scale poses. Since the self-supervised model \cite{kumar2020fisheyedistancenet,kumar2021fisheyedistancenet++,kumar2021svdistnet} for fisheye images has not been open-sourced and its employed WoodScape \cite{yogamani2019woodscape} dataset does not provide ground-truth depth for fisheye images, we did not include it in our experiments.

Our FisheyeDepth model achieves the best results across all evaluation metrics. This is because other monocular depth estimation models do not specifically address the distortion and aberration present in fisheye images, whereas our proposed fisheye projection ensures consistency in depth estimation. Notably, the use of real-scale poses during training in the FSNet model also improves depth estimation performance, further demonstrating that using true poses is more robust than poses estimated by PoseNet.

\subsection{Visualization Results in KITTI-360}
Fig. \ref{result360} presents the visualization results of depth estimation from the KITTI-360 dataset. We conduct experiments using full-size fisheye images with a 180-degree FOV, which exhibit significant distortion and aberration, especially in the peripheral regions. Despite these challenges, our FisheyeDepth model accurately estimates the depth.

All the visualizations are derived from urban road scenes, with image backgrounds that include a variety of elements such as roads, vegetation, buildings, and cars. Notably, in images (b), (c), and (l), shadow occlusions did not affect the model's ability to accurately estimate the depth of black cars. Additionally, in images (h) and (m), the model successfully estimated the depth of the pole and tunnel. The overall visualization results demonstrate the model's robustness in depth estimation across varying distances, from near to far.

\begin{figure}[t!]
    \centering
    \includegraphics[width=.5\textwidth]{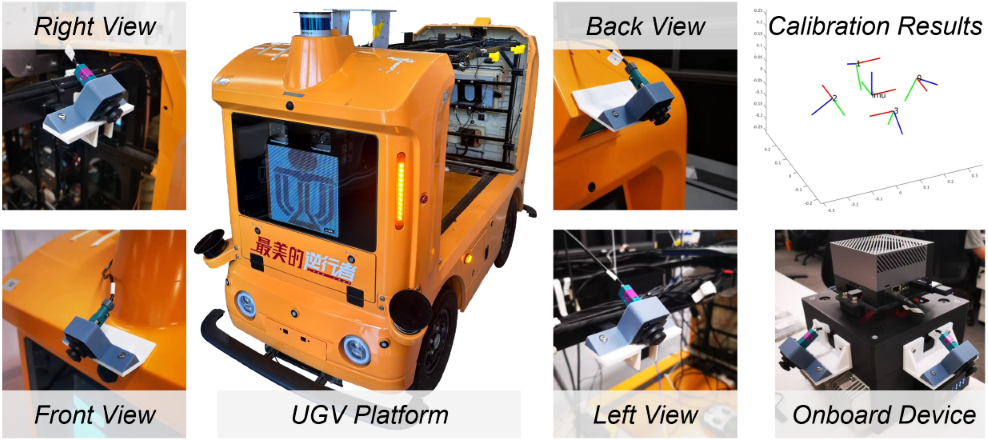}
    \vspace{-15pt}
    \caption{\textbf{Setup of the real scene experiment.} We use four calibrated fisheye cameras on the UGV platform for data collection.}
    \label{exp-set}
    \vspace{-15pt}
\end{figure}

\begin{table*}[t]
\renewcommand\arraystretch{1.45}
\caption{Ablation experiments on different strategies of FisheyeDepth model. (FFT means flash fisheye projection, Real Scale means training using the robot’s pose, and MC means multi-channel output.)}
\vspace{-8pt}
\centering
\footnotesize
\begin{tabular}{>{\centering\arraybackslash}m{2.1cm}cccccccccc}
\hline 
Methods & FFT & Real Scale & MC & \cellcolor{cyan!30}Abs Rel\(\downarrow\) & \cellcolor{cyan!30}Sq Rel\(\downarrow\) & \cellcolor{cyan!30}RMSE\(\downarrow\) & \cellcolor{cyan!30}RMSE$_{\log }$\(\downarrow\) & \cellcolor{pink!30}$\delta<1.25$ \(\uparrow\) & \cellcolor{pink!30}$\delta<1.25^2$ \(\uparrow\) & \cellcolor{pink!30}$\delta<1.25^3$ \(\uparrow\) \\
\hline 
\bf{Monodepth2} \cite{godard2019digging}  & - & - & - & 0.602 & 1.804 & 1.788 &  0.563 & 0.426 & 0.685 & 0.835 \\
\hline
\multirow{4}{*}{\bf{FisheyeDepth}} 
 & \textcolor{green}{\ding{51}} & \textcolor{red}{\ding{55}} & \textcolor{red}{\ding{55}} & 0.205 & 0.547 & 1.066 & 0.266 & 0.726  & 0.903 & 0.971 \\
 & \textcolor{green}{\ding{51}} & \textcolor{green}{\ding{51}} & \textcolor{red}{\ding{55}} & 0.115  & 0.672 & 1.163 & 0.173 & 0.912 & 0.972 & 0.987 \\
 & \textcolor{green}{\ding{51}} & \textcolor{red}{\ding{55}} & \textcolor{green}{\ding{51}} & 0.126 & 0.647 & 1.150  & 0.180 & 0.893 & 0.969  & 0.987 \\
 & \textcolor{green}{\ding{51}} & \textcolor{green}{\ding{51}} & \textcolor{green}{\ding{51}} & \textbf{0.106} & \textbf{0.407} & \textbf{0.927} & \textbf{0.170} & \textbf{0.920} & \textbf{0.975} & \textbf{0.989} \\
\hline
\end{tabular}
\label{Table2}
\vspace{-5pt}
\end{table*}

\begin{figure*}[t!]
    \centering
    \includegraphics[width=.99\textwidth]{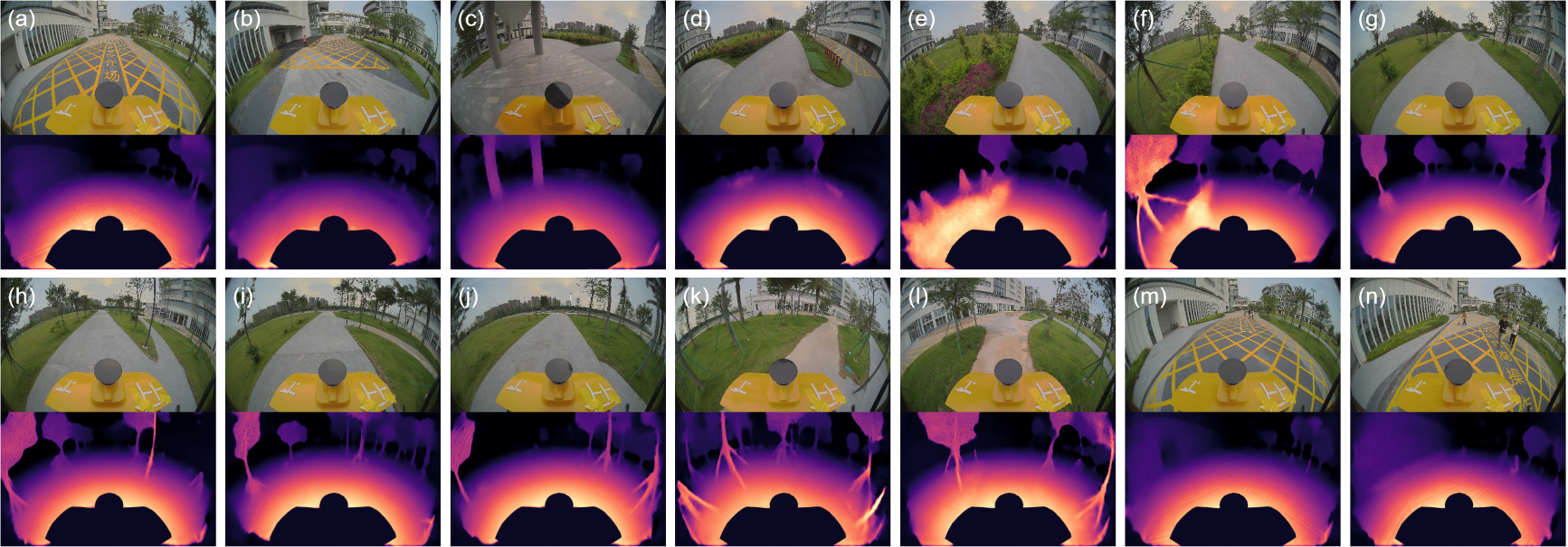}
    \vspace{-8pt}
    \caption{\textbf{Depth estimation visualization in the real scene.} The image background contains complex roads, narrow passages, and different obstacles.
}
    \label{result-real}
    \vspace{-15pt}
\end{figure*}

\subsection{Ablation Study}

We conducted ablation experiments to analyze the effects of different strategies within the FisheyeDepth model, with the results shown in Table \ref{Table2}. The main focus was on testing the impact of projection of the fisheye camera model, the training with real scale poses, and the multi-channel output on the model's performance.


Among these, the improvement in depth estimation accuracy from the camera model projection is the most significant, as it effectively eliminates the distortion in fisheye image projections, ensuring consistency in depth predictions. This is the primary reason why the FisheyeDepth model can perform stable depth estimation.

Incorporating training with real-scale poses also significantly enhanced the depth estimation performance, as the robot's odometry provides stable pose variations.
The multi-channel output module improves performance by optimizing and integrating depth predictions at different stages at the feature level. Importantly, it reduces the impact of real pose noise on training through optimization, thereby jointly promoting robust depth estimation of the FisheyeDepth model in real-scale robotic scenarios.

\subsection{Real Scene Experiment}

Since our FisheyeDepth model is a self-supervised training approach capable of stable training without ground-truth depth, we conducted real-world experiments in road scenes at HKUST Guangzhou campus. The experimental platform setup is shown in Fig. \ref{exp-set}. We used an Unmanned Ground Vehicle (UGV) as the robotic platform, equipped with four calibrated fisheye cameras mounted on the onboard device for data collection and model inference.

We jointly trained on images from the four fisheye cameras to enhance the model's generalization to different scenes, and then performed visual analyses on the most representative front-facing fisheye images. The robot's real poses used during training were provided by visual inertial odometry. The visualization results of the experiments are presented in Fig. \ref{result-real}. The experiments encompassed various campus scenes, including roads with complex backgrounds, narrow pathways, and environments with nearby obstacles.

Notably, in images (a), (m), and (n), the backgrounds are filled with cluttered yellow lines, yet this did not lead to incorrect depth estimations of the road. The FisheyeDepth model was able to accurately estimate the depth of trees with indistinct trunk features, especially in images (k) and (l), where the scenes are surrounded by multiple slender trees. Overall, without any ground-truth depth assistance, we achieved robust depth estimation on fisheye images through self-supervised training using only the robot's real poses.


\section{CONCLUSIONS}


In this paper, we present FisheyeDepth, a true-scale self-supervised depth estimation model designed specifically for fisheye cameras. By incorporating a fisheye camera model to handle image projective distortion and integrating real-scale pose information from sensor fusion, we eliminate scale ambiguity and improve depth prediction accuracy and stability. Additionally, our multi-channel output module enhances robustness through adaptive feature fusion, making the model highly effective for robotic and autonomous navigation tasks.

Extensive experiments on the KITTI-360 dataset and real-world scenarios demonstrate the superior performance of our model compared to monocular self-supervised depth estimation methods. The results confirm that integrating real-scale poses and addressing fisheye distortions are crucial for achieving high-precision depth estimation, especially in challenging outdoor environments.

Our work provides a scalable and practical framework for real-world applications, particularly in autonomous driving and robotics. Future research may explore the potential to extend this model to other types of omnidirectional cameras.





\clearpage

\bibliographystyle{ieeetr} 
\bibliography{ref}         

\end{document}